\documentclass[nohyperref]{article}

\usepackage{microtype}
\usepackage{graphicx}
\usepackage{subfigure}
\usepackage{booktabs} 
\graphicspath{ {./img/} }

\usepackage[backref=page]{hyperref}

\usepackage[accepted]{icml2022}

\usepackage{amsmath}
\usepackage{amssymb}
\usepackage{mathtools}
\usepackage{amsthm}
\usepackage{commath}

\usepackage[capitalize,noabbrev]{cleveref}

\usepackage{xcolor}
\definecolor{cbrown}{RGB}{177, 89, 40}
\definecolor{cgreen}{RGB}{51, 160, 44.17} 

\newcommand{\sqboxs}{1.2ex}
\newcommand{\sqbox}[1]{\textcolor{#1}{\rule{\sqboxs}{\sqboxs}}}

\theoremstyle{plain}
\newtheorem{theorem}{Theorem}[section]

\theoremstyle{definition}
\newtheorem{definition}[theorem]{Definition}

\theoremstyle{remark}

\usepackage[textsize=tiny]{todonotes}

\newcommand{\e}{\varepsilon}

\newcommand{\agt}{\texttt{agt}}
\newcommand{\dev}{\texttt{dev}}

\newcommand{\lauro}[1]{\textcolor{blue}{\textbf{lauro}: #1}}
\newcommand{\edouard}[1]{\textcolor{green}{\textbf{edouard}: #1}}

\renewcommand{\lauro}[1]{}
\renewcommand{\edouard}[1]{}

\icmltitlerunning{Goal Misgeneralization in Deep Reinforcement Learning}

\begin{document}

\twocolumn[
\icmltitle{Goal Misgeneralization in Deep Reinforcement Learning}

\icmlsetsymbol{equal}{*}

\begin{icmlauthorlist}
\icmlauthor{Lauro Langosco}{equal,cam}
\icmlauthor{Jack Koch}{equal}
\icmlauthor{Lee Sharkey}{equal,tub}
\icmlauthor{Jacob Pfau}{edi}
\icmlauthor{Laurent Orseau}{dm} 
\icmlauthor{David Krueger}{cam}
\end{icmlauthorlist}

\icmlaffiliation{cam}{University of Cambridge}
\icmlaffiliation{tub}{University of Tübingen}
\icmlaffiliation{edi}{University of Edinburgh}
\icmlaffiliation{dm}{DeepMind, London}

\icmlcorrespondingauthor{Lauro Langosco}{langosco.lauro@gmail.com}

\icmlkeywords{Machine Learning, ICML, Reinforcement Learning, RL, AI Safety, AI Alignment, Alignment, Robustness, Generalization, Misgeneralization, Safety, Inner Alignment, Objective Robustness, Goal Misgeneralization}

\vskip 0.3in
]



\printAffiliationsAndNotice{\icmlEqualContribution} 

\begin{abstract}
We study \emph{goal misgeneralization}, a type of out-of-distribution generalization failure in reinforcement learning (RL). 
Goal misgeneralization occurs when an RL agent retains its capabilities out-of-distribution yet pursues the wrong goal. 
For instance, an agent might continue to competently avoid obstacles, but navigate to the wrong place.
In contrast, previous works have typically focused on capability generalization failures, where an agent fails to do anything sensible at test time.
We formalize this distinction between capability and goal generalization, provide the first empirical demonstrations of goal misgeneralization, and present a partial characterization of its causes.
\end{abstract}

\section{Introduction} \label{sec:intro}
Out-of-distribution (OOD) generalization, performing well on test data that is not distributed identically to the training set, is a fundamental problem in machine learning \citep{ood-in-ml}.
OOD generalization is crucial since in many applications it is not feasible to collect data distributed identically to that which the model will encounter in deployment.

\begin{figure}\label{fig:coinrun}
 \centering
 \includegraphics[width=\columnwidth]{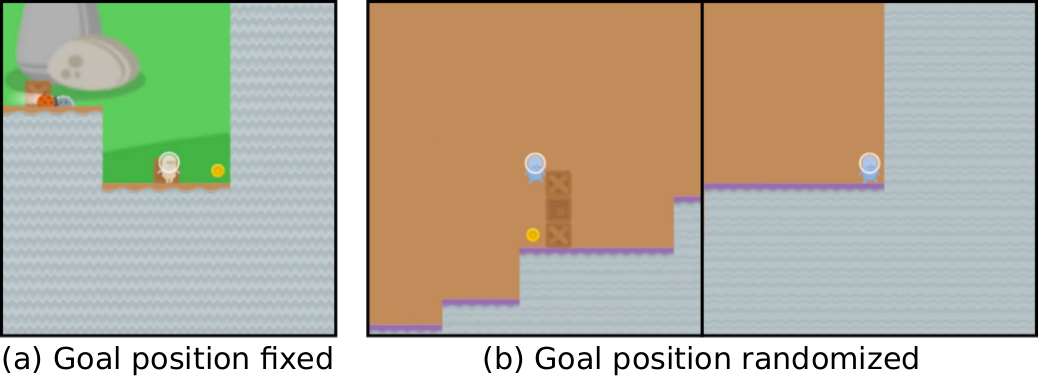}
  \caption{
\textbf{(a)} At training time, the agent learns to reliably reach the coin which is always located at the end of the level. \textbf{(b)} However, when the coin position is randomized at test time, the agent still goes towards the end of the level and often skips the coin. The agent's capability for solving the levels generalizes, but its goal of collecting coins does not.
}
\end{figure}

In this work, we focus on a particularly concerning type of generalization failure that can occur in RL.
When an RL agent is deployed out of distribution, it may simply fail to take useful actions. However, there exists
an alternative failure mode in which the agent pursues a goal other than the training reward while retaining the capabilities it had on the training distribution.
For example, an agent trained to pursue a fixed coin might not recognize the coin when it is positioned elsewhere, and instead competently navigate to the wrong position (Figure~\ref{fig:coinrun}).
We call this kind of failure \textbf{goal misgeneralization}\footnote{
  We adopt this term from \citet{goalMisgenDeepMind}. A previous version of our work used the term `objective robustness failure' instead. We use the term `goal' to refer to goal-directed (optimizing) behavior, \emph{not} just goal-states in MDPs.
} and distinguish it from \textbf{capability generalization} failures. 
We provide the first empirical demonstrations of goal misgeneralization to highlight and illustrate this phenomenon.

While it is well-known that the true reward function can be unidentifiable in inverse reinforcement learning \citep{unidentifiability}, our work shows that a similar problem can also occur in reinforcement learning when features of the environment are correlated and predictive of the reward on the training distribution but not OOD.
In this way, goal misgeneralization can also resemble problems that arise in supervised learning when models use unreliable features: both problems are a form of competent misgeneralization that works in-distribution but fails OOD.
Disentangling capability and goal generalization failures is difficult in supervised learning; for instance, are adversarial examples bugs or features \citep{ilyas}?
In contrast, studying RL allows us to formally distinguish capabilities and goals, which roughly correspond to understanding the environment dynamics and the reward function, respectively.

Goal misgeneralization might be more dangerous than capability generalization failures, since an agent that capably pursues an incorrect goal can leverage its capabilities to visit arbitrarily bad states \citep{zhuang2021consequences}.
In contrast, the only risks from capability generalization failures are those of accidents due to incompetence.

An important implication of goal misgeneralization is that
training a model by optimizing an objective $R$ is not enough to guarantee that the model will itself learn to pursue $R$ rather than some proxy for $R$.
    This is especially concerning in the context of \emph{AGI safety}: ensuring that advanced AI systems are safe despite being capable enough to escape our control \citep{bostrom_superintelligence}.
Systems that competently pursue a misaligned goal may tend to seek power and deceive their operators for instrumental reasons \citep{turner_optimal, omohundro}. 
With highly advanced AI systems, this could lead to human disempowerment: for example, an AI system might prevent its operators from shutting it down \citep{hadfield2017off, soares}. 
If complex tasks are rife with proxies for their training objectives, it may be very hard to predict what objectives the trained AI systems will have \citep{mesa-optimizers}.

Our main contributions are:
\vspace{-3mm}
\begin{itemize}
\setlength\itemsep{0.1em}
  \item We formalize goal misgeneralization, distinguishing it from capability generalization failures 
  (Section~\ref{sec:or}), and experimentally validate our definition on a gridworld environment (Section~\ref{sec:agt-dev}).
  \item We experimentally showcase goal misgeneralization.
  Specifically, deep RL agents trained on the Procgen benchmark \citep{procgen}---a set of diverse procedurally generated environments specifically designed to induce robust generalization---still fail on our slightly modified environments (Section~\ref{sec:experiments}).
  \item We show that goal misgeneralization may be alleviated by increasing the diversity of the training distribution so that the agent learns to distinguish the reward from proxies (Sections~\ref{subsec:pct-rand} and \ref{sec:maze}).
  \item We investigate the causes of goal misgeneralization.  In particular, we find that the actor and the critic components of our actor-critic model learn different proxy goals (Section~\ref{sec:actor-critic}).
\end{itemize}

\section{Goal Misgeneralization} \label{sec:or}

Goal misgeneralization is a type of OOD generalization failure.
OOD generalization is usually studied in the supervised learning setting, where it is defined as achieving good test performance on data sampled from a distribution other than the training distribution. 
%
We focus on the reinforcement learning setting \citep{sutton}, in which a system is trained to take actions in an environment in order to maximize a given reward. 
In this setting, the problem is to achieve high reward despite a shift in the distribution of observations or the transition dynamics. 
OOD generalization problems frequently arise in RL and are an active area of research \citep{kirk2021}.
However, goal misgeneralization in particular has not been the focus of any previous academic work.
Studying this class of failures is particularly important from the point of view of machine learning safety \citep{hendrycks2021unsolved}, since agents that pursue imperfect proxies may fail suddenly \citep{alex2022effects, ibarz2018reward} and catastrophically \citep{zhuang2021consequences} as their capabilities increase.
With this in mind, we provide a definition of goal misgeneralization and show how it can be formalized.

\subsection{Defining Goal Misgeneralization}
A deep RL agent is trained to maximize a reward $R \colon S \times A \times S \to \mathbb R$, where $S$ and $A$ are the sets of all valid states and actions, respectively. Assume that the agent is deployed out-of-distribution; that is, an aspect of the environment (and therefore the distribution of observations) changes at test time. 
\textbf{Goal misgeneralization} occurs if the agent now achieves low reward in the new environment because it continues to act capably yet appears to optimize a different reward $R' \neq R$.
\edouard{Not sure how far down this rabbit hole you want to go, but an agent's actions will of course always be compatible with multiple reward functions (see, e.g., Ng \& Russell's Policy Invariance under Reward Transformations, or DeepMind's EPIC). So the more precise definition might be to say something like: the policy the agent is \emph{compatible} with some set $\mathcal{R}'$ of reward functions, while the \emph{assigned} reward function $R \not \in \mathcal{R}'$.}
We call $R$ the \textbf{intended objective} and $R'$ the \textbf{behavioral objective} of the agent.

Formally, we follow \citet{orseau2018agents} in distinguishing goal-directed policies (\emph{agents}) from unoptimized policies (\emph{devices}). 
Let $\eta_\agt(R)$ and $\eta_\dev(d)$ be priors over a space of reward functions $R \in \mathcal R$ and a space of devices (policies) $d \in \Pi$ respectively. 
Further let $p_\agt(\tau \mid R)$ and $p_\dev(\tau \mid d)$ be
the likelihood functions giving
the probability of a trajectory $\tau$ given a particular objective $R$ or device $d$.
We define two distributions over trajectories, the agent mixture $p_\texttt{agt}$ and the device mixture $p_\texttt{dev}$:
\begin{align} \label{eq:agt-dev}
    p_\agt(\tau) &= \sum_{R \in \mathcal R} p_\agt(\tau \mid R)\ \eta_\agt(R),\\
    p_\dev(\tau) &= \sum_{d \in \Pi} p_\dev(\tau \mid d)\ \eta_\dev(d).
\end{align}

The choice of device likelihood $p_\dev(\tau \mid d)$ is straightforward: we simply choose the distribution over trajectories induced by running the policy $d$ in the environment.
For the agent likelihood $p_\agt(\tau \mid R)$, a popular choice is the maximum entropy model 
$p_\agt(\tau \mid R) \propto \exp(R(\tau))$ 
\citep{ziebart2008maximum}. Another possibility is to choose $p_\agt(\tau \mid R)$ to be the probability density of the random trajectory obtained by training an RL algorithm to maximize $R$ and collecting rollouts.\footnote{This requires an RL algorithm and model (e.g.\ neural network). In practice, this choice of $p(\tau \mid R)$ is intractable to compute.}

\begin{definition}[Goal misgeneralization] \label{def:or}
A policy $\pi$ undergoes \emph{goal misgeneralization} if test reward is low and $p_\texttt{agt}(\tau) > p_\texttt{dev}(\tau)$ holds on average for the trajectories induced by $\pi$ in the OOD test environment. In other words, the policy is acting in a goal-directed manner, but not achieving high reward.
We can infer a posterior distribution over behavioral objectives:
\begin{equation*}
\eta_\agt(R \mid \tau) \propto p_\agt(\tau \mid R) \eta_\agt(R).
\end{equation*}
\end{definition}

In Section~\ref{sec:agt-dev} we compute these mixtures explicitly and validate Definition~\ref{def:or} in a gridworld environment.


\subsection{Causes of Goal Misgeneralization} \label{subsec:causes}
When should we expect models to learn robust goals? We begin by suggesting possible prerequisites for goal misgeneralization:
\vspace{-3mm}
\begin{enumerate}
\setlength\itemsep{0.1em}
  \item The training environment must be diverse enough to learn sufficiently robust capabilities. 
  \item There must exist some proxy $R': S \times A \times S \to \mathbb R$ that correlates with the intended objective on the training distribution, but comes apart (i.e.\ is much less correlated, or anti-correlated) on the OOD test environment.
\end{enumerate}
These conditions are necessary for goal misgeneralization to arise: If (1) is not the case, then RL algorithms tend to memorize simple action sequences that work in the training environment but are not robust under distributional shift \citep{procgen}. Meanwhile (2) is necessary because by assumption the policy achieves high (training) reward; thus the behavioral objective must be correlated with the intended objective on the training environment. However, (1) and (2) are by no means sufficient since, by themselves, they do not guarantee that the model learns to pursue the proxy reward $R'$ instead of the intended objective.

We note that assumptions (1) and (2) are quite weak: almost every real-world problem requires a diverse training environment (to learn robust capabilities), and proxies are common in complex environments. 
Thus goal misgeneralization depends mostly on whether the inductive biases of the model and training algorithm prime it to learn a proxy that then diverges from the intended objective on the test set.
%
We expect that learned proxies will: 
\vspace{-2mm}
\begin{itemize}
\setlength\itemsep{0.1em}
  \item be correlated with the intended objective $R$ on the training distribution but not necessarily the test distribution.
  \item tend to be easier to learn than the intended objective $R$ because a proxy $R'$ may: 

    \vspace{-1.5mm}
    \begin{itemize}
    \setlength\itemsep{0.1em}
    \item use features that are simpler or more favored by the inductive biases of the model compared with the intended objective \citep{valleperez2019deep, geirhos2020}.
    \item be denser than the intended objective \citep{singh2010intrinsically}. 
    \end{itemize}
\end{itemize}

For example, despite being a product of evolution (which optimizes for genetic fitness), humans tend to be more concerned with proxy goals, such as food or love, than with maximizing the number of their descendants. 
This illustrates a general phenomenon: given a challenging goal (such as ``maximize fitness''), complex environments are rife with proxies and sub-goals (such as ``eat rich food'') of that goal, many of which are more dense or simpler to optimize than the original goal. 
This observation has previously been made by \citet{singh2010intrinsically}, who also draw the analogy with evolution, and note that bounded agents (i.e.\ with limited experience and/or computation) will often achieve higher expected reward \textit{according to the true reward} when trained to optimize a proxy reward function.


\section{Experiments} \label{sec:experiments}
Having defined goal misgeneralization and outlined when and why we expect it to occur, 
we now present experiments 
designed to demonstrate different kinds of goal misgeneralization and distinguish them from capability generalization failures.\footnote{
Our code can be found at
\url{https://github.com/JacobPfau/procgenAISC} (Environments) and
\url{https://github.com/jbkjr/train-procgen-pytorch} (Training).

Video examples of goal misgeneralization in all of the following environments can be found    \href{https://drive.google.com/drive/folders/17d2wzn7nI0Yl_TcCNOsoVg9EvnZoKHfN?usp=sharing}{at this link}.
}

In each experiment, we train an agent that performs capably when deployed out-of-distribution, but pursues a behavioral objective different from the objective for which it was trained. This behavior is consistent across multiple random seeds for training.

For each of our experiments we hypothesize a behavioral objective that the policy has learned: navigating to the right-hand end of the level (CoinRun), navigating to the upper right corner (Maze I), navigating to the yellow object (Maze II) and gathering keys (Keys and Chests). None of these is a robust proxy for the intended objective. It is possible that there exist alternate objectives that also explain this behavior: for example, navigating towards a tall, left-facing wall (CoinRun). 
For our purposes, it is enough to show that a plausible proxy objective exists. 
Nonetheless, we conduct a series of experiments that confirm the `move right' hypothesis over the `move to wall' hypothesis for the CoinRun agent's behavioral objective (see Section~\ref{sec:actor-critic}).

\begin{figure}[t]
\vskip 0.2in
\begin{center}
\centerline{\includegraphics[width=7.5cm]{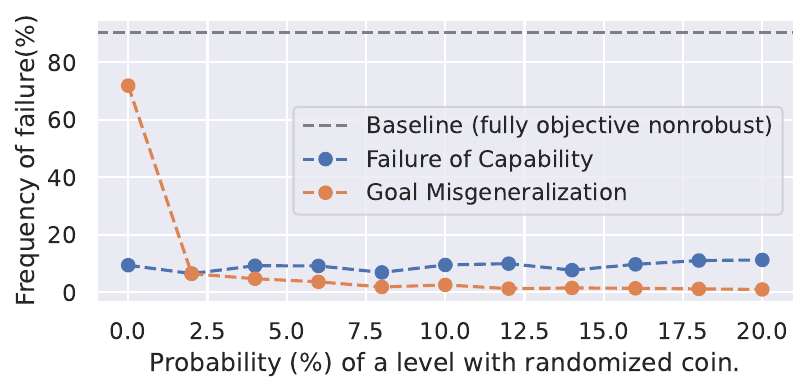}}
\caption{
  Goal generalization is greatly improved in our CoinRun experiments when just 2\% of training levels have randomly placed coins (like the test environment). 
  More randomization helps.
  Baseline: Since even a policy that entirely ignores the coin may sometimes hit the coin by accident, we compute a base rate for a `fully misgeneralizing' policy that treats the coin as invisible.
}
\label{fig:ablations}
\end{center}
\vskip -0.2in
\end{figure}

We follow a zero-shot protocol in all experiments except Figure~\ref{fig:ablations}: the agent does not see the (OOD) testing environment during training.
Except in Section~\ref{sec:agt-dev}, all environments are adapted from the Procgen environment suite \citep{procgen}. This suite is built to study sample efficiency and generalization to within-distribution tasks. Agents (feedforward neural networks trained using Proximal Policy Optimization - further details in the Appendix) are tasked with performing well in an arcade-like video game from pixel observations. The environments are procedurally generated and thus diverse; to perform well, an agent must learn strategies that work in a wide range of task settings and difficulties and cannot rely on e.g.\ memorizing a small number of trajectories to solve a fixed set of levels.
This diversity alone is insufficient to prevent goal misgeneralization, however; diversity of a different sort is needed, as we demonstrate in Figure~\ref{fig:ablations}.

\vspace{-2mm}\paragraph{Different kinds of  failure.} The experiments illustrate different flavors of goal misgeneralization.
\emph{Directional proxies} (CoinRun): the agent learns to move to the right instead of to the true source of reward (the coin).
\emph{Location  proxies} (CoinRun, Maze I): In Maze I, the agent learns to navigate to the upper right corner instead of to the true source of reward (the cheese). The critic---but not the actor---also learns such a proxy in CoinRun.
\emph{Observation ambiguity} (Maze II): The observations contain multiple features that identify the goal state, which come apart in the OOD test distribution.
\emph{Instrumental goals} (Keys and Chests): The agent learns an objective (collecting keys) that is only instrumentally useful to acquiring the intended objective (opening chests).

\subsection{CoinRun} \label{sec:coinrun}

In the Procgen CoinRun environment, the agent spawns on the left side of the level and has to avoid enemies and obstacles to get to a coin.
The coin yields a reward of 10, all other rewards are 0.
In our training environments, the coin is always located at the right end of the level next to a wall; reaching the coin terminates the episode.
To evaluate goal misgeneralization, we create test environments in which the coin is located in a random (accessible) location.

After training, the agent competently navigates to the end of the level in the training environment.
At test time, the agent generally ignores the coin completely and proceeds to the end of the level, as shown in Figure~\ref{fig:coinrun}.
This suggests that the agent has learned the proxy objective of ``move right'' rather than ``move to the coin''.
It competently achieves this behavioral objective, which is perfectly correlated with the intended objective on the training distribution and appears to be easier for the agent to learn; nevertheless, test reward is low because the behavioral objective misgeneralizes.

\begin{figure}[ht]
 \centering
 \includegraphics[width=7cm]{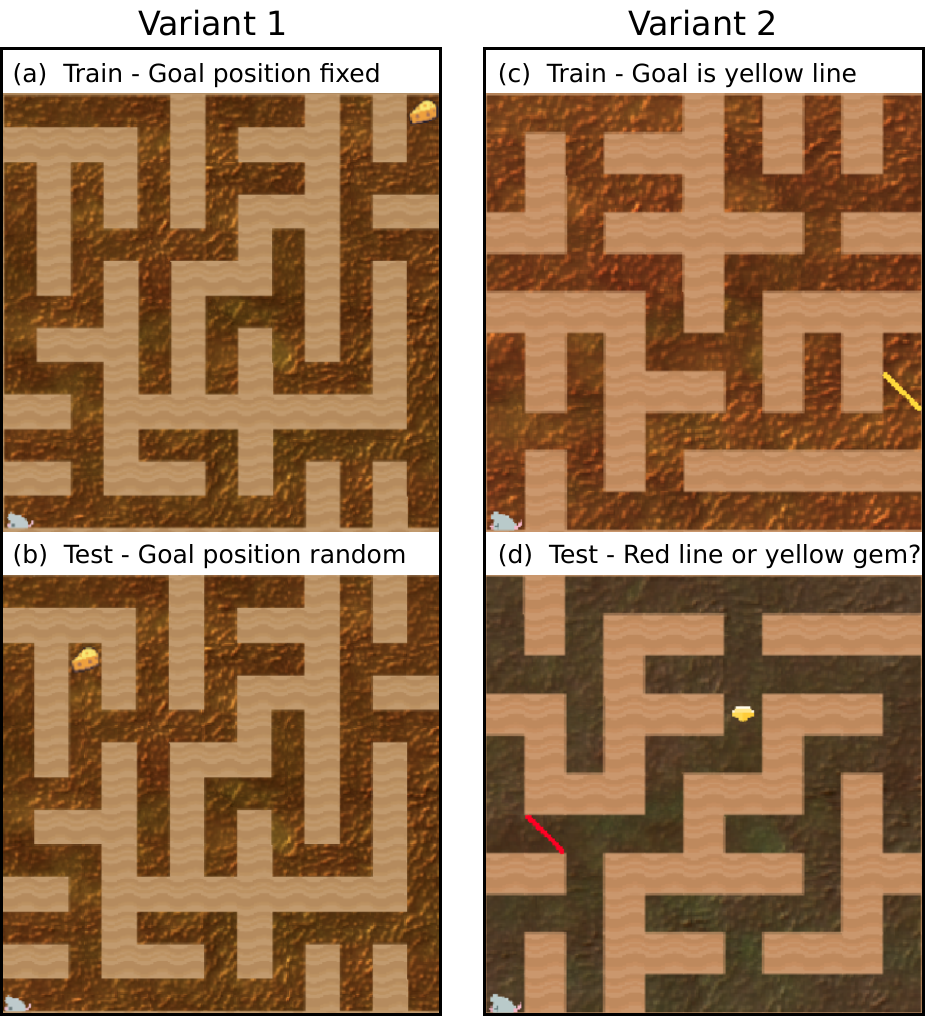}
  \caption{
  The agent (the mouse) is trained to navigate mazes to reach its goal.
  \textbf{(a \& b)} An agent is trained on procedurally-generated mazes with the cheese in a fixed position (top right corner) ignores it and navigates to the top right corner when the cheese is placed randomly.
  \textbf{(c \& d)} An agent trained to navigate to a yellow line consistently navigates to a yellow gem when deployed in environments in which there are only red lines and yellow gems.
  If it is meant to collect lines and not gems, this is a case of goal misgeneralization.
  }\label{fig:maze}
\end{figure}

\vspace{-2mm}\paragraph{Training with randomly placed coins.}
\label{subsec:pct-rand}
To test how consistent goal misgeneralization is, we train a series of agents on environments which vary in how often the coin is placed randomly.
Results can be seen in Figure~\ref{fig:ablations}, which shows the frequencies of two different outcomes:
\vspace{-3mm}
\begin{enumerate}
\setlength\itemsep{0.1em}
    \item \textbf{Failure of capability:} the agent dies or gets stuck, thus neither getting the coin nor to the end of the level.  This is evaluated on the training environments where the coin is typically at the end of the level.
    \item \textbf{Goal misgeneralization:} the agent misses the coin and navigates to the end of the level.  This is evaluated on the OOD test environments where coin location is randomized.
\end{enumerate}
As expected, as the diversity of the training environment increases, the probability of goal misgeneralization decreases, as the model learns to pursue the coin instead of going to the end of the level. 
We also include a baseline which measures the rate at which an invisible ``coin'' would be captured, to determine how often the coin would be captured by an agent that completely ignores it.
We see that even when the coin is always at the end of the level during training, the rate of goal misgeneralization is lower than this baseline.




\subsection{Maze} \label{sec:maze}

\paragraph{Variant 1.} \label{subsec:mazeI}
We modify the Procgen Maze environment in order to implement an idea from \citet{hubinger-investigation}. 
In the original environment, a maze is generated using Kruskal's algorithm \citep{kruskal}, and the agent is trained to navigate towards a piece of cheese located at a random spot in the maze.

\begin{figure}[t]
 \centering
 \includegraphics{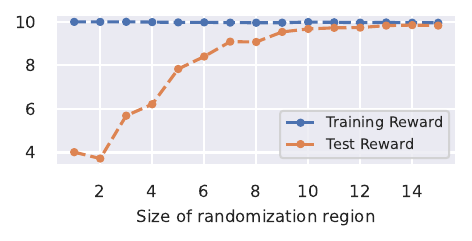}
  \vspace{-3mm}
  \caption{
  The goal is randomly located within a region of size $1-16$ in the upper right corner of the maze. As the region grows, validation performance on the fully randomized environment improves (i.e.\ correct goal generalization is more likely).  However, the agent still uses location as a proxy until the region is quite large.
    }
\label{fig:maze-sweep}
\end{figure}

We modify the original environment so that the cheese is always in the upper right corner (Figure~\ref{fig:maze}a). As in the CoinRun experiment, when an agent is trained on the environment with a consistent reward location but tested in an environment with a random reward location, the agent ignores the randomly placed objective, instead navigating to the upper right corner of the maze (Figure~\ref{fig:maze}b). The intended objective is to reach the cheese, but the behavioral objective of the learned policy is to navigate to the upper right corner.  
Somewhat surprisingly, we also find that the agent continues to pursue a proxy objective of ``move to the upper right corner'' even when this proxy becomes imperfect (see Figure~\ref{fig:maze-sweep}).

\paragraph{Variant 2.} \label{subsec:mazeII}
In the experiments so far, goal misgeneralization arises due to an ambiguity between a visual feature (coin / cheese) and a positional feature (right / top right) which come apart at test time.
To illustrate a different kind of distributional shift, we present a simple setting in which there is no \emph{positional} feature that favors one objective over the other; instead, the agent is forced to choose between two ambiguous visual cues.

We train an RL agent on a version of the Procgen Maze environment where the reward is a randomly placed \emph{yellow diagonal line} (Figure~\ref{fig:maze}c). At test time, we deploy it on a modified environment featuring two randomly placed objects: a yellow \emph{gem} and a \emph{red} diagonal line; the agent is forced to choose between consistency in shape or in color (Figure~\ref{fig:maze}d).
Except for occasionally getting stuck in a corner, the agent usually pursues the yellow gem, thus generalizing in favor of color rather than shape consistency (89\% of the time, excluding occasions where it must pass through the red line to get to the yellow gem, $n=102$).
As in previous examples, training with the correct reward function is not enough to guarantee correct goal generalization here; rather, another approach such as increasing environment diversity or using a different inductive bias may be necessary to specify the intended OOD behavior. 


\subsection{Keys and Chests} \label{subsec:keys-and-chests}

\begin{figure}[t]
 \centering
 \includegraphics[width=\columnwidth]{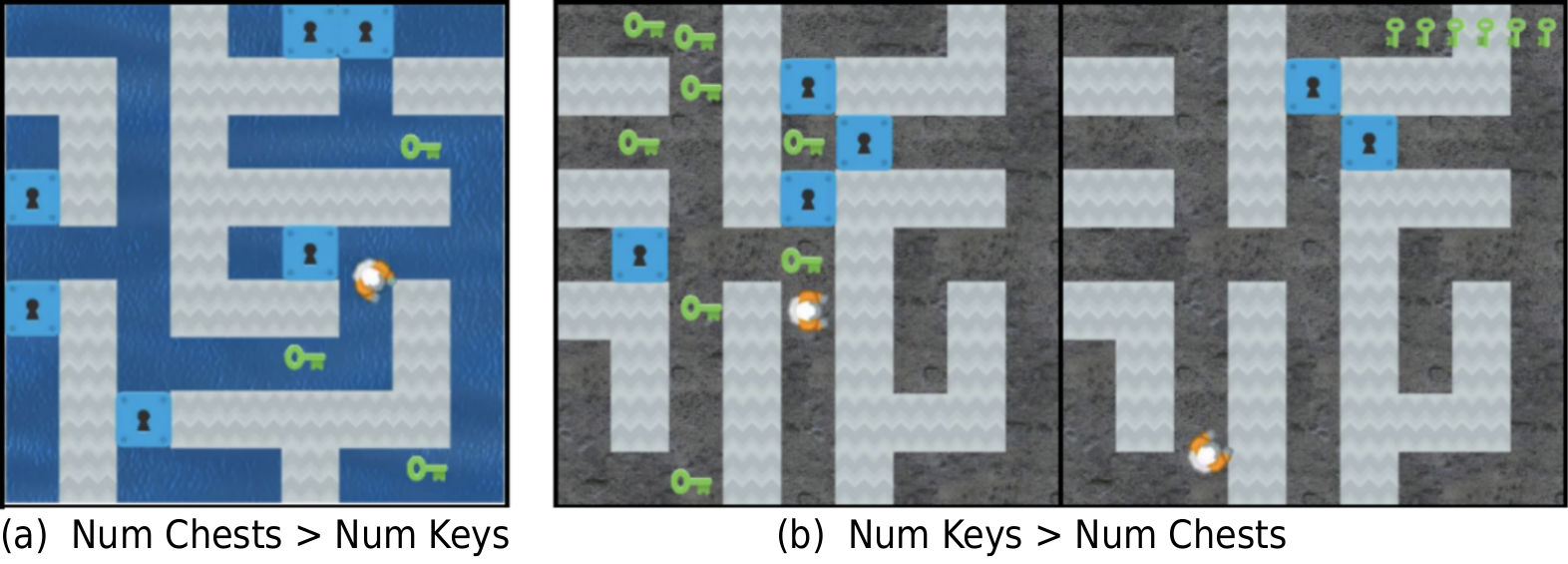}
  \caption{Goal misgeneralization on the ``Keys and chests" task. 
  The agent must collect keys in order to open chests and is only rewarded for opening chests. 
  \textbf{(a)} The agent is trained on procedurally-generated mazes in which there are twice as many chests as keys.
  \textbf{(b)} At test time, there are instead twice as many keys as chests. The agent routinely prioritizes collecting all the keys before opening any remaining chests despite the fact that doing so offers no benefit to its episode reward (in fact, it would \emph{decrease} its time-discounted return).
  }\label{fig:keyschests}
\end{figure}

So far, our experiments featured environments in which there was a 
proxy that was perfectly correlated 
with the intended objective on the training distribution. 
The Keys and Chests environment, first suggested by \citet{keys-and-chests}, provides a different type of example. This environment, which we implement by adapting the Heist environment from Procgen, is a maze with two kinds of objects: keys and chests. 
Whenever the agent comes across a key it is added to a key inventory. 
When an agent with at least one key in its inventory comes across a chest, the chest is opened and a key is deleted from the inventory. 
The agent is rewarded for every chest it opens.

As in previous experiments, we induce goal misgeneralization by subjecting the agent to different training and test environment distributions: In the training environment, there are twice as many chests as keys, while in the test environment there are twice as many keys as chests. The basic task facing the agent is the same (the reward is only given upon opening a chest), but the circumstances are different.

We observe that an agent trained on the ``many chests'' distribution goes out of its way to collect all the keys before opening the last chest on the ``many keys'' distribution (Figure~\ref{fig:keyschests} and Figure~\ref{fig:kc-training}, Appendix), even though only half of them are even instrumentally useful for the intended objective; occasionally, it even gets distracted by the keys in the inventory (which are displayed in the top right corner) and spends the rest of the episode trying to collect them instead of opening the remaining chest(s).

We describe the agent as having learned a simple behavioral objective: collect as many keys as possible, while sometimes visiting chests. This strategy leads to high reward in an environment where chests are plentiful and the agent can thus focus on looking for keys. 
One reason that the agent may have learned this proxy is that the proxy is less sparse than the intended objective while nevertheless being correlated with it on the training distribution. However, the proxy fails when keys are plentiful and chests are no longer easily available.

\subsection{Critic Generalization vs. Actor-Critic Generalization}
\label{sec:actor-critic}
All of the experiments above use PPO \citep{ppo}, an actor-critic method \citep{sutton1998reinforcement}.
In these methods, the policy (``actor'') learns to optimize an approximate value function provided by the ``critic''.
So far, we've demonstrated goal misgeneralization, where the actor behaves in a goal-directed manner but doesn't achieve high test reward. 
In this section we analyze the CoinRun experiment more closely and show that the actor and the critic \emph{both} fail to generalize OOD; furthermore, they \emph{fail in different ways}. We conclude that the actor and the critic have different inductive biases that lead them to fail in different ways.

\begin{figure}
 \centering
 \includegraphics[width=\columnwidth]{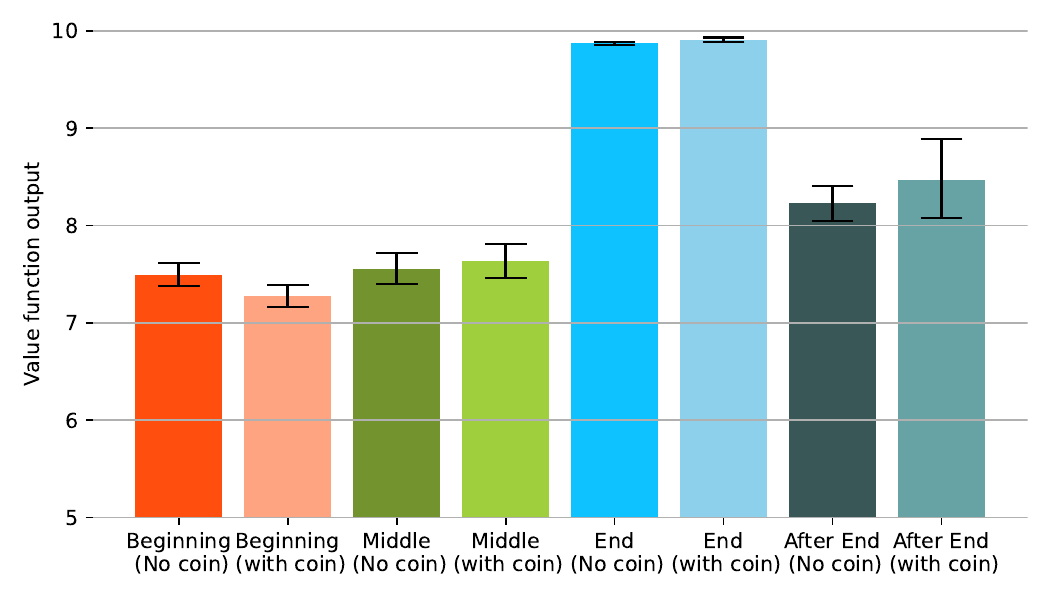}
  \caption{The average value function output for images ($n=950$) at different stages of CoinRun levels, with and without a coin visible. Error bars are bootstrapped 95\% confidence intervals. The coin has an insignificant effect at all stages of a level.
  }\label{fig:value_bar_plot}
\end{figure}

\paragraph{Critic Misgeneralization.}
In order to determine how much the critic values the coin (the intended objective) vs.\ reaching the end of the level (the proxy objective), we compare the value it assigns to states where these factors are varied (Figure~\ref{fig:value_bar_plot}).
We find that the value (i.e.\ the output of the critic) is much higher at the end of the level than elsewhere, and that
the presence of the coin makes no discernible difference.
Thus we conclude that the critic misgeneralizes, assigning high value to the proxy instead of the intended objective.
To help identify the features in observations at the end of the level that cause higher value, we generate attribution maps by taking the gradient of the value function output with respect to the observation, following \citet{simonyan2013deep}. The end-wall is highlighted at least as much as the coin (Figure~\ref{fig:attribution_maps}, Appendix).

\paragraph{Actor-Critic Inconsistency.}

\begin{figure*}
 \centering
 \includegraphics[width=2\columnwidth]{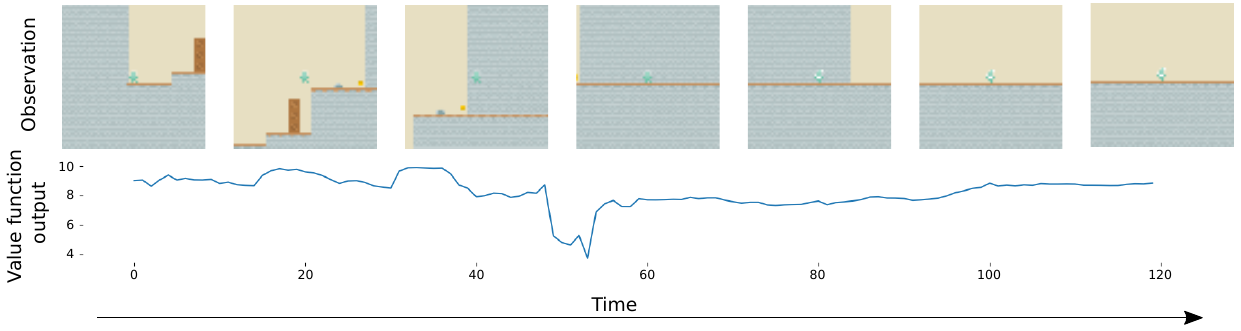}
\caption{Observations and critic's value estimate for a typical OOD episode with permeable end wall. The agent continues to move to the right, through the wall. This happens even though the critic assigns the highest value around timestep 35, when the agent is just left of the wall (where the coin is typically located during training). This phenomenon occurs $100\%$ of the time that the agent reaches the permeable wall (n=114). 
This demonstrates that the actor's behavioural proxy objective differs from the critic's proxy objective. Such differences could cause goal misgeneralization even in situations where a critic has learned the true value function.
}
  \label{fig:permeable_wall}
\end{figure*}

In Section \ref{sec:coinrun} we established that the actor (the policy) misgeneralizes, and in the previous paragraph we have shown that the critic also misgeneralizes. Here we show that the behavior of the actor and the output of the critic are in fact inconsistent: the actor navigates as far right as possible even when this involves moving past a wall, whereas the critic assigns highest value to states in which the agent is touching a wall before having moved past it.
We deploy the agent in an environment with a permeable end wall. 
If the actor generalized consistently with respect to the critic, it should stay at the wall, or return to it upon passing through it.
Instead, whenever the agent reaches the end wall it continues moving right and passes through the wall $100\%$ of the  time ($n=114$) (Figure~\ref{fig:permeable_wall}). 
This indicates that the policy pursues a ``move right'' proxy objective, rather than the ``move to the wall'' proxy objective of the critic, or the intended ``move to the coin'' objective.
In other words, the actor learns a ``non-robust proxy of a non-robust proxy''. 
Its failure to match the critic's proxy objective is another source of and example of goal misgeneralization.

\subsection{Measuring Agency} \label{sec:agt-dev} 
\begin{figure}[ht]
\begin{center}
\centerline{\includegraphics[width=0.8\columnwidth]{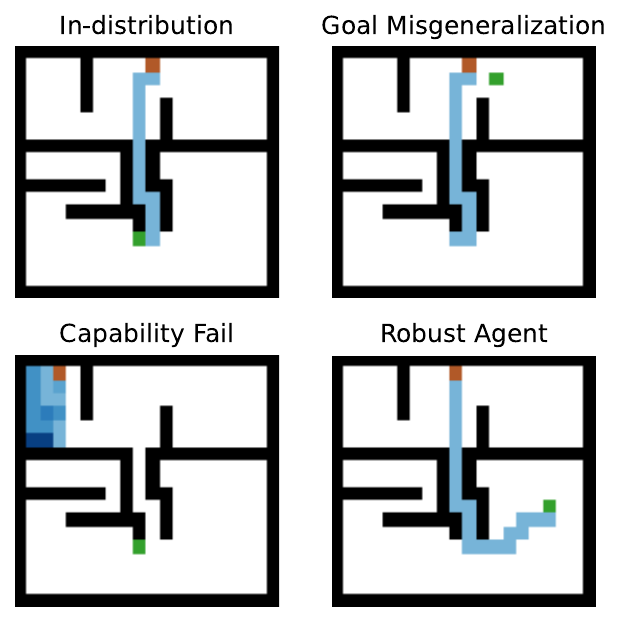}}
\caption{
\sqbox{cbrown} Start.
\sqbox{cgreen} Goal.
\emph{In-distribution:} the policy reaches its (fixed) goal.
\emph{Goal misgeneralization}: the policy navigates to the wrong position when the goal is moved. \emph{Capability Generalization Failure:} when start position is moved, the policy gets stuck. \emph{Robust:} the policy always reaches the goal for all start / goal positions.
}
\label{fig:agt-dev}
\end{center}
\end{figure}

We validate the formal definition of goal misgeneralization from Section~\ref{sec:or} by explicitly computing the agent and device mixtures in a gridworld environment based on work by \citet{orseau2018agents}, shown in Figure~\ref{fig:agt-dev}.
In this environment there are $4$ possible actions (move up, down, left, right). The 
state consists of 
two sets of (row, column) coordinates: the position of the agent and of the goal. Possible goal states include every accessible square in the gridworld; formally, our set of possible reward functions is
\[
\mathcal R = \{R_{s} \mid s \in S\},
\]
where $S$ is the set of accessible squares in the gridworld and $R_{s}(s') = 1$ if $s = s'$ and $0$ otherwise.

We generate trajectories of an agent attempting to reach a goal cell $g$. We distinguish four types of trajectories (Figure~\ref{fig:agt-dev}); depending on the type, the goal position is either random or fixed.
We distinguish capability from goal generalization failure by comparing the mixture probabilities $p_\texttt{agt}(\tau)$ and $p_\texttt{dev}(\tau)$ (Table~\ref{tab:agt-dev}).
A detailed description of the trajectory types and the computation of the mixture probabilities is available in Appendix~\ref{app:agt-dev}.



Consider a policy that successfully solves a maze in which the locations of the start state and goal state are fixed (Figure~\ref{fig:agt-dev}, top left).
There are three ways this policy might generalize OOD, illustrated in Figure~\ref{fig:agt-dev}. 
\begin{enumerate}
    \item A goal misgeneralizing policy might reliably navigate to the location where the goal was during training, ignoring its actual location (Figure~\ref{fig:agt-dev}, top right).
    \item A policy that fails at capability generalization might memorize the trajectory from start to goal, and behave randomly on other states (Figure~\ref{fig:agt-dev}, bottom left).
    \item A robust policy would reliably solve the task for any location of goal and start state (Figure~\ref{fig:agt-dev}, bottom right).
\end{enumerate}

As shown in Table~\ref{tab:agt-dev}, the agents \& devices formalism successfully distinguishes goal misgeneralization from capability generalization failures: The robust policy as well as the misgeneralizing policy are clearly recognized as goal-directed agents, whereas the policy that fails at capability generalization is correctly classified as non-agent.

\begin{center}
\begin{table}[!h]
\begin{tabular}{|c | c c c|} 
 \hline
 Type & $-\log p_\texttt{agt}(\tau)$ & $-\log p_\texttt{dev}(\tau)$ & $p(\texttt{agt} \mid \tau)$ \\ [0.5ex] 
 \hline \hline
 IID & 5.7 & 20.8 & 0.9999 \\ 
 \hline
 G.\ Misg. & 14.1 & 30 & 0.9999 \\
 \hline
 Cap.\ Fail & 72 & 69 & 0.0674 \\
 \hline
 Robust & 10.5 & 30.5 & 0.9999 \\
 \hline
\end{tabular}
\caption{As expected, all trajectories from Figure~\ref{fig:agt-dev} are classified as agents, except the capability generalization failure (Cap.\ Fail).}
\label{tab:agt-dev}
\end{table}
\end{center}

\section{Related Work} \label{sec:related}

\paragraph{Out-of-Distribution Generalization.}
Goal misgeneralization is a form of out-of-distribution (OOD) generalization failure.
OOD generalization is the problem of performing well on a novel distribution at test time.
Causes for such a train-test mismatch include: i) the training data does not characterize the true distribution \citep{dataset-bias}, ii) the distribution shifts over time \citep{dataset-shift}, and iii) the test data are adversarially perturbed \citep{szegedy, goodfellow2015explaining}. 
OOD generalization is a well established limitation of existing deep learning approaches, which can be very sensitive to changes in distribution \citep{recht2019imagenet, imagenetc}, and may base their predictions on shortcuts \citep{geirhos2020} or spurious correlations \citep{beery, irm}.
Such lack of robustness may be due to underspecification \citep{damour2020underspecification}: since there are many patterns a deep network can learn which explain the training distribution equally well, practitioners may need to provide additional information to disambiguate between these possible solutions.
Underspecification of the mapping from state to reward is present in our CoinRun and Maze environments, making goal misgeneralization \textit{unavoidable} if the inductive biases of the deep RL algorithms deployed don't match the intended behavior.
The existing work on OOD generalization is largely complementary to our work here on goal misgeneralization.
For example, learning invariant predictors \citep{irm, rex} across diverse training environments might help a model learn the true cause of the reward it receives and improve goal misgeneralization.
We also add to existing work on OOD generalization by highlighting that when a model fails to generalize OOD, it may do so in two different ways that have notably different consequences: it might generalize completely incapably, or it might generalize capably but pursue an incorrect objective.
This distinction is important because pursuing an incorrect objective can lead to different---and potentially more severe---consequences \citep{zhuang2021consequences}.
OOD generalization is especially important in online RL because updating the policy leads to a shift in the training distribution.

\vspace{-2mm}\paragraph{Generalization in RL.}
We define and study goal misgeneralization in the context of reinforcement learning.
Historically, generalization in RL received little attention, but many recent works address various forms of RL generalization, including OOD generalization.  
Notable directions of research include sim-to-real \citep{peng2018sim}, robust RL \citep{morimoto2005robust}, and offline RL \citep{levine2020offline}; see \citet{kirk2021} for a review.
Solving classic deep RL environments such as ATARI \citep{bellemare2013arcade} may already require generalizing across states, but \citet{procgen} note that overfitting to a particular environment is commonly observed, and propose diverse sets of environments to promote generalization.
While \citet{procgen} use the same distribution of environments during training and test time, we modify their environments to create OOD test environments.


\vspace{-2mm}\paragraph{Goal Misgeneralization / Objective Robustness.} An earlier public version of this work used the term `objective robustness failure' in the place of `goal misgeneralization'. We adopt the term `goal misgeneralization' from \citet{goalMisgenDeepMind} in order to avoid confusion with reward misspecification and to avoid having two terms for the same phenomenon.
Previous work on OOD generalization has largely failed to distinguish between goal misgeneralization and capability generalization failures.
\citet{mesa-optimizers} and \citet{2d-robustness} are perhaps the first to make such a distinction explicitly, and the term \emph{objective robustness failure} is used by \citet{clarifying} to refer to the former failure mode. 
These works also argue that goal misgeneralization may be catastrophic, motivating our focus on this type of failure.
Previously, \citet{Leike} used the term \emph{reward-result gap} to refer to the difference between what a model was optimiz\emph{ed} for and what it appears to be optimiz\emph{ing} (i.e.\ what we call the behavioral objective).
We add to these works by formalizing the distinction between capability generalization failure and goal misgeneralization, and providing the first empirical demonstrations of goal misgeneralization.

\vspace{-2mm}\paragraph{Mesa-Optimization.}
Public non-academic discussions of concerns related to goal misgeneralization, and the analogy with evolution described in Section~\ref{subsec:causes}, go back at least to 2016 \citep{yudkowsky_2016, christiano_2016}.\footnote{
Terms used in these discussions include ``subsystem reasoning'' \citep{taylor_2017}, ``optimization daemons'', ``inner optimizers'',  and ``inner alignment'' \citep{lesswrong_mesaopt}.}
These early discussions, as well as \citet{mesa-optimizers}, focus on goal misgeneralization caused by \textbf{mesa-optimization}, a phenomenon where a model learns an optimization process (even if not explicitly trained to do so).
Mesa-optimization could lead to goal misgeneralization if the learned ``inner objective'' optimized differs from the ``outer objective'' specified by the designer, but this need not be the case.
Furthermore, goal misgeneralization can occur without mesa-optimization.
Thus these are in fact two distinct behaviors, and our work does not demonstrate or address mesa-optimization.\footnote{For a sufficiently broad definition of mesa-optimization, goal misgeneralization may become equivalent to misaligned mesa-optimization. We use a different term to emphasize the connection to OOD generalization and not depend on a notion of optimization \citep{shah2021}.}
Mesa-optimization could be a concern independent of goal misgeneralization if the mesa-optimizer pursues undesirable \textit{means} of optimizing the correct objective \citep{krueger_maharaj_leike}, e.g.\ we might not want a prediction system to make self-fulfilling prophecies \citep{armstrong_oracle}.
Furthermore, while we've defined goal misgeneralization as a form of OOD failure, mesa-optimization may lead to undesirable behavior such as deception \citep{lesswrong_mesaopt} or power-seeking \citep{turner_optimal} on-distribution.







\vspace{-2mm}\paragraph{Unidentifiability in Inverse Reinforcement Learning.}
\edouard{The EPIC paper might also be worth citing in this section.} 
Goal misgeneralization tends to arise when there are multiple possible reward functions that are indistinguishable from the intended objective and produce similar behavior on the training set, but not OOD. 
This type of unidentifiability is analogous to the one encountered in inverse reinforcement learning (IRL).
\citet{unidentifiability} separate the causes for this unidentifiability in IRL into two classes.
The first, \textbf{representational unidentifiability}, arises because some transformations of reward functions, e.g.\ rescaling, preserve the \textit{relative} returns of different policies. 
The second, \textbf{experimental unidentifiability}, occurs when $\pi$'s observed behavior is optimal under two (or more) reward functions which are not functionally equivalent---i.e.\ there exist situations where they would entail different optimal behavior.
Goal misgeneralization can arise from experimental unidentifiability when an agent only encounters situations that distinguish its behavioral objective from the intended objective function at test time.


\vspace{-2mm}\paragraph{Reward Misspecification.}
Reward specification is the problem of specifying a reward that captures the behavior we want \citep{concrete-problems,faulty-reward}. Goal misgeneralization is a distinct problem: it may lead to failure even if the reward function is perfectly specified.\footnote{Failures due to reward misspecification occur when the model behaves in an unintended way that nevertheless scores highly on the reward function. In contrast, in goal misgeneralization, models score \emph{poorly} on the training reward because they are pursuing a different objective.}
%
Reward misspecification can produce similar failures to goal misgeneralization, however, when the designer specifies a proxy objective that yields good training performance, but fails OOD \citep{hadfieldmenell2017inverse}. 

\section{Discussion} \label{sec:discussion}
We have formally defined the problem of goal misgeneralization in RL, and provided the first explicit examples of goal misgeneralization in deep RL systems.
We argue that goal misgeneralization is a natural category since, much like adversarial robustness failures, goal misgeneralization has distinct causes and poses distinct problems.


Our definition of goal misgeneralization via the agent and device mixtures is practically limited: it is generally hard to define a useful prior over objectives, and the computation quickly becomes intractable for large and complex environments. Conceptually, the division into agents and devices is somewhat restrictive; for example, multi-agent systems do not naturally fit into the framework.

Better understanding agency and optimization remains an important avenue for future work. There is a number of interesting questions in this direction, such as formalizing how some part of the world can optimize some other part of the world and thus be an agent \emph{embedded} in its environment~\citep{demski2019embedded}, and understanding when deep learning systems are likely to behave like agents optimizing proxy objectives.

Future empirical work may also study the factors that influence goal misgeneralization. For instance, what kinds of proxy objectives are agents most likely to learn? 
This may help us understand what kinds of environment diversity are most useful for learning robust goals.

\section{Contributions}
JK and LL independently proposed the idea of demonstrating goal misgeneralization. LS suggested to use Procgen for experiments and conceived of the CoinRun demonstration; JK, LL, LS, and JP set up and trained the agent on CoinRun. LL and JP modified the Procgen environments, and LL ran the sweeps in CoinRun and Maze. LS with assistance from LL conceived and ran the experiments in section 3.4. LS ran the attribution map experiments.
DK became involved after the original arXiv preprint; he proposed defining goal misgeneralization via agents and devices \cite{orseau2018agents}, proposed the experiment in Figure 4, and made major contributions to the writing and presentation. 
Laurent Orseau\footnote{Laurent joined and contributed after the ICML deadline for author inclusion, which is why he is included as author here but not on the official ICML submission.}
ran the `measuring agency' experiment (Section~\ref{sec:agt-dev}), following a specification designed by LL and DK.
The manuscript was written by DK, JK, LL, and LS.

\section*{Acknowledgements}

Special thanks to Rohin Shah and Evan Hubinger for their guidance and feedback throughout the course of this project, and to Rohin for proposing the term \emph{goal misgeneralization}. Thanks also to Max Chiswick for assistance adapting the code for training the agents, Adam Gleave, Dan Hendrycks, Edouard Harris, Robert Kirk, and Dmitrii Krasheninnikov for helpful feedback on drafts of this paper, and the organizers of the AI Safety Camp for bringing the authors of this paper together: Remmelt Ellen, Nicholas Goldowsky-Dill, Rebecca Baron, Max Chiswick, and Richard Möhn.

This work was supported by funding from the AI Safety Camp and Open Philanthropy. Lee Sharkey was supported by the Centre for Effective Altruism Long Term Future Fund and by the Deutsche Forschungsgemeinschaft (DFG, German Research Foundation) under Germany's Excellence Strategy–EXC2064/1–390727645.

\newpage
\bibliography{refs}
\bibliographystyle{icml2022}

\newpage
\onecolumn
\appendix

\section{Implementation details} \label{sec:implementation}

For all environments, we use an Actor-Critic architecture using Proximal Policy Optimization (PPO) \citep{ppo}. 
The architecture is based on the architecture used in \cite{espeholt2018impala} but omits the recurrent components of the original network. Both the actor (policy function) and critic (value function) are implemented by feedforward neural networks on top of a shared residual convolutional network. 
All models are implemented in PyTorch \citep{pytorch} and our implementations are based on a codebase by \citet{pg-pt}. 
Unless otherwise stated, models are trained on $100$k procedurally generated levels for 200M timesteps. We use the Adam optimizer \citep{adam} in all experiments.
Each training run required approximately 30 GPU hours of compute on a V100.

\section{Experiment Details}
\subsection{Measuring Agency} \label{app:agt-dev}
\subsubsection{Generating Trajectories}
Fix positions $s$ and $g$ (`start' and `goal') in a 20x20 gridworld. Then we generate trajectories $\tau_1, \dots, \tau_n$ as follows. 
For every trajectory $\tau_i$, we sample random positions $s_\text{rand}^{(i)}$ and $g_\text{rand}^{(i)}$. Note that $s_\text{rand}^{(i)}$ and $g_\text{rand}^{(i)}$ take on new values for every trajectory, while $s$ and $g$ are fixed. For every trajectory, we also identify one of the gridworld states as the \emph{intended goal} $g_\text{true}^{(i)}$. We then generate four types of trajectories $\tau_i^1, \tau_i^2, \tau_i^3, \tau_i^4$:

\begin{enumerate}
    \item Set $g_\text{true}^{(i)} = g$. Pick the trajectory that takes the shortest path from $s_\text{rand}^{(i)}$ to $g_\text{true}^{(i)}$ (`In-distribution').
    \item Set $g_\text{true}^{(i)} = g_\text{rand}^{(i)}$. Pick the trajectory that takes the shortest path from $s_\text{rand}^{(i)}$ to $g$ (`Goal Misgeneralization'),
    \item Set $g_\text{true}^{(i)} = g$. Pick the trajectory that starts at $s_\text{rand}^{(i)}$ and moves in a uniformly random direction every step, for $50$ timesteps. If the trajectory ever crosses the shortest path from $s$ to $g$, then it follows that path to $g$ (`Capability Failure').
    \item Set $g_\text{true}^{(i)} = g_\text{rand}^{(i)}$. Pick the trajectory that takes the shortest path from $s_\text{rand}^{(i)}$ to $g_\text{true}^{(i)}$ (`Robust agent').
\end{enumerate}
We are left with $4n$ trajectories $(\tau_i^\lambda)_{i \leq n, \lambda \leq 4}$. Note:
\begin{enumerate}
    \item The trajectories $\tau_i^1$ are generated from a policy that can reach the fixed goal state $g_\text{true}^{(i)} = g$ from any place on the grid.
    \item The trajectories $\tau_i^2$ are generated from the same policy, deployed in an environment where the goal state $g_\text{true}^{(i)}$ is changed. The policy still navigates to the fixed position $g$, but this is no longer the correct goal; this behavior is designed to match the behavior we saw in the policies we trained for the Maze experiments in Section~\ref{subsec:mazeI}.
    \item The trajectories $\tau_i^3$ are designed to imitate the capability generalization failure of a policy which navigates from a fixed start state to a fixed end state. When initialized to a random start location, the policy takes random actions since it only knows to navigate along a fixed path.
    \item The trajectories $\tau_i^4$ are generated from a policy that robustly takes the shortest path to $g_\text{true}^{(i)}$ from any position in the gridworld even when the goal state is randomized.
\end{enumerate}

\subsubsection{Calculating Mixture Probabilities}
We follow the method in \citet{orseau2018agents}. The observations available to agent policies include the goal state and the position of the agent.

\paragraph{Agent prior.} We specify the set of possible goal states to consist of all $n^2$ locations in the gridworld. (Those familiar with \citet{orseau2018agents} should note that we do not use the switching prior).

\paragraph{Agent mixture.} We specify the set of goals to consist of all accessible squares in the gridworld, plus the (variable) goal $g_\text{true}^{(i)}$. Note that $g_\text{true}$ can be random in the cases where we set $g_\text{true}^{(i)} = g_\text{rand}^{(i)}$), and thus vary from trajectory to trajectory.
Formally, our set of objectives is 
$$\mathcal R = \{R_{s} \mid s \in S\cup \{g_\text{true}\} \},$$
where $S$ is the set of accessible squares in the gridworld and $R_{s}(s') = 1$ if $s = s'$ and $0$ otherwise. We then take a uniform prior $\eta_\agt(R) = 1 / \vert \mathcal R \vert$ over this set.
Given an objective $R$, define the probability $p_\varepsilon(\tau \mid R)$ of a trajectory as induced by an $\varepsilon$-greedy policy.
Here, the observations of the policy consist of the (row, column) position of the agent. 
We then integrate over $\e$:
\begin{equation*}
    p_\agt(\tau \mid R) = \int_0^1 p_\e(\tau \mid R) \dif \e.
\end{equation*}

\paragraph{Device mixture.}
Recall that a device is just a stochastic, tabular policy that takes in an observation and outputs an action. The observation consists of the type of cell (empty, wall, start, goal) that the device is facing, in the direction of its last action.
Our device prior $\eta_\dev$ is uniform over the space of policies. 
Set $p_\e(\tau \mid d)$ to be the probability of a trajectory generated by acting in an $\e$-deterministic way with respect to $d$, that is take the action determined by $d$ with probability $1-\e$ and a random action otherwise. Just as previously we integrate over $\e$ in $[0, 1]$ to compute the final likelihood $p_\dev(\tau \mid d)$.

\section{Hyperparameters} \label{sec:hyperparams}

\begin{table}[h!]
\centering
\caption{Hyperparameters}
\begin{tabular}{lcc}
\hline Hyperparameter & Value \\
\hline
ENV. DISTRIBUTION MODE & HARD \\
$\gamma$ & $.999$ \\
$\lambda$ & $.95$ \\
LEARNING RATE & $0.0005$ \\
\# TIMESTEPS PER ROLLOUT & $256$ \\
EPOCHS PER ROLLOUT & $3$ \\
\# MINIBATCHES PER EPOCH & $8$ \\
MINIBATCH SIZE & $2048$ \\
ENTROPY BONUS $\left(k_{H}\right)$ & $.01$ \\
PPO CLIP RANGE & $.2$ \\
REWARD NORMALIZATION? & YES \\
LEARNING RATE & $5 \times 10^{-4}$ \\
\# WORKERS & 4  \\
\# ENVIRONMENTS PER WORKER & 64 \\
TOTAL TIMESTEPS & $200 \mathrm{M}$ \\
ARCHITECTURE & Impala \\
LSTM? & No  \\
FRAME STACK? & No \\
\hline
\end{tabular}
\end{table}

\section{Value attribution maps and other figures} \label{sec:value_attr_maps_suppl}
\begin{figure}
 \centering
 \includegraphics[width=12.0cm]{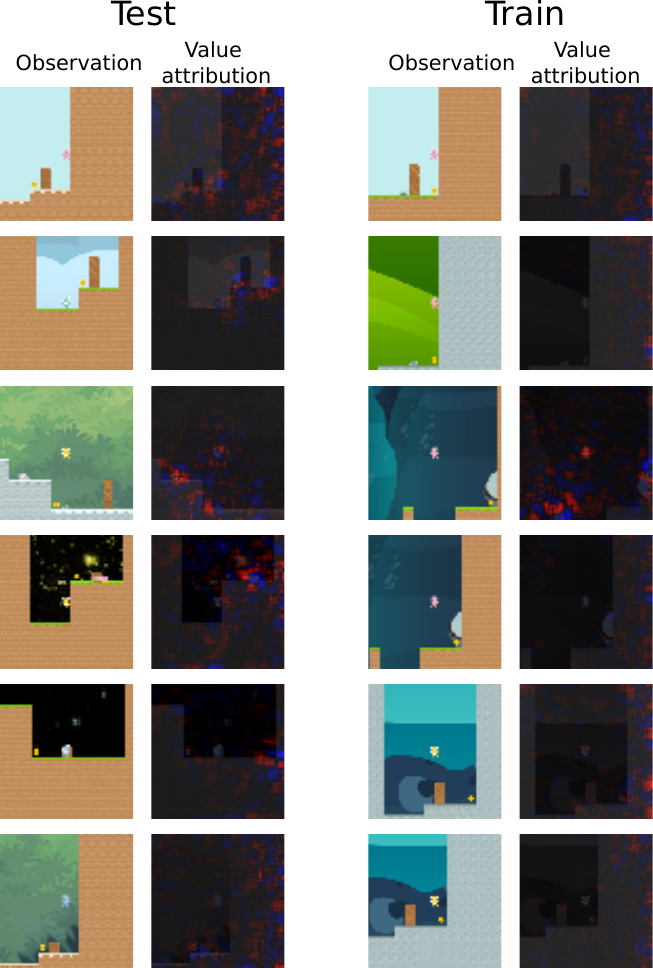}
  \caption{Attribution maps of the agent's observation with respect to its value function output. Maps were generated by taking the gradient of the value function output with respect to the observation pixels (averaged over channels) \citep{simonyan2013deep}. 
  Red shading indicates pixels that negatively influence value function output and blue shading indicates pixels that have positive influence. The pixel level attributions were standardized by dividing each map by the value of the largest absolute magnitude of pixel attribution. The attribution maps are passed through a Gaussian blur transform with kernel size 5 and $\sigma=5$. As observed in \citet{hilton2020understanding}, we find that the sign of the attribution map is often difficult to understand - for instance, buzzsaws might sometimes appear to have positive attribution rather than negative. We therefore focus on the absolute magnitude of the attribution. In both the training and test environment, the agent's value function assigns large attribution to the end wall and occasionally the coin, enemies, and buzzsaws. From the attribution plots alone, we can only determine that the end wall appears more important to the agent than the coin, but the coin might nevertheless also be somewhat important for the value function output.}\label{fig:attribution_maps}
\end{figure}

\begin{figure}[ht]
\begin{center}
\centerline{\includegraphics[width=9.5cm]{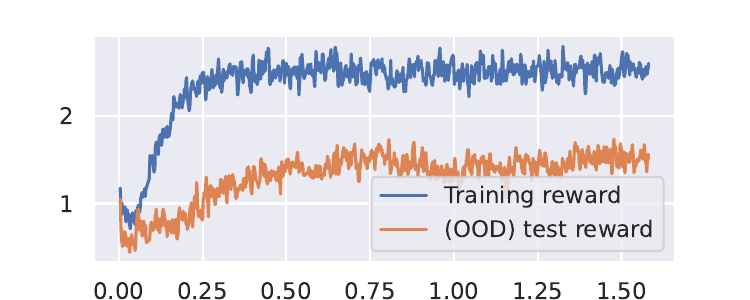}}
\caption{
  Average return during training of the Keys \& Chests agent. The reward on the `many keys' test environment is much lower than the `many chests' training reward.
}
\label{fig:kc-training}
\end{center}
\end{figure}

\begin{figure}[ht]
\begin{center}
\centerline{\includegraphics[width=9.5cm]{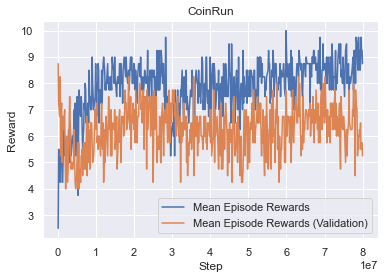}}
\caption{
  Expected return during training of the CoinRun agent.
}
\label{fig:cr-training}
\end{center}
\end{figure}

\begin{figure}[ht]
\begin{center}
\centerline{\includegraphics[width=9.5cm]{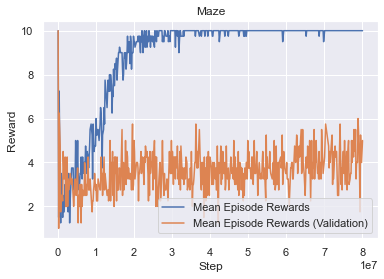}}
\caption{
  Expected return during training of the maze agent.
}
\label{fig:maze-training}
\end{center}
\end{figure}

\end{document}